\documentclass{article}

\usepackage{authblk}
\usepackage{times}                      % Use Times New Roman font
\usepackage[a4paper, margin=1in]{geometry} % Set page size and margins (consolidated)
\usepackage{abstract}                   % For abstract layout control
\usepackage{multicol}
\usepackage{marvosym}

\usepackage{amsmath}                    % For advanced math environments
\usepackage{amssymb}                    % For math symbols like \checkmark
\usepackage{pifont}                     % For symbols like \ding{55}

\usepackage{graphicx}                   % To include figures
\usepackage{booktabs}                   % For professional table lines (\toprule, etc.)
\usepackage{tabularx}                   % For tables with auto-wrapping text columns
\usepackage{ragged2e}                   % For better text alignment in tables (\RaggedRight)
\usepackage{caption}                    % For customizing captions
\usepackage{float}                      % The modern replacement for 'here' to provide [H]
\newsavebox{\myboxtable}

\usepackage{algorithm}
\usepackage{algpseudocode}

\usepackage{natbib}                      % For handling citations
\usepackage[hidelinks]{hyperref}        % For clickable links 

\title{\textbf{Uncertainty-Aware Generative Oversampling Using an Entropy-Guided Conditional Variational Autoencoder}}

\author{Amirhossein Zare$^1$, Amirhessam Zare$^1$, Parmida Sadat Pezeshki$^1$, 
Herlock (SeyedAbolfazl) Rahimi$^2$, Ali Ebrahimi$^3$, 
Ignacio Vázquez-García$^{4,5}$, Leo Anthony Celi$^{6,7,8}$}

\affil{\small
$^1$School of Medicine, Tehran University of Medical Sciences, Tehran, Iran \quad
$^2$Department of Electrical and Computer Engineering, Yale University, New Haven, CT, USA \quad
$^3$SDU Health Informatics and Technology, The Maersk Mc-Kinney Moller Institute, University of Southern Denmark, Odense, Denmark \quad
$^4$Department of Pathology and Krantz Family Center for Cancer Research, Massachusetts General Hospital, Harvard Medical School, Boston, MA, USA \quad
$^5$Broad Institute of MIT and Harvard, Cambridge, MA, USA \quad
$^6$Laboratory for Computational Physiology, MIT, Cambridge, MA, USA \quad
$^7$Division of Pulmonary, Critical Care \& Sleep Medicine, Beth Israel Deaconess Medical Center, Boston, MA, USA \quad
$^8$Department of Biostatistics, Harvard T.H. Chan School of Public Health, Boston, MA, USA}

\date{} 

\begin{document}

% --- RENDER TITLE AND ABSTRACT (FULL WIDTH) ---
\maketitle
% ---FOOTNOTE BLOCK ---
{\renewcommand{\thefootnote}{}
\footnote{\Letter~ \texttt{amhosseinzare@gmail.com}, \texttt{amir.hessam.zare@gmail.com}, \texttt{pezeshkiparmida@gmail.com}, \texttt{herlock.rahimi@yale.edu}, \texttt{aleb@mmmi.sdu.dk}, \texttt{ivazquez-garcia@mgh.harvard.edu}, \texttt{leoanthonyceli@yahoo.com}}
\footnote{Source code available at: \url{https://github.com/Amirhossein-Zare/LEO-CVAE}}}

% ---------------------------
\begin{abstract}
\noindent 
Class imbalance remains a major challenge in machine learning, especially for high-dimensional biomedical data where nonlinear manifold structures dominate. Traditional oversampling methods such as SMOTE rely on local linear interpolation, often producing implausible synthetic samples. Deep generative models like Conditional Variational Autoencoders (CVAEs) better capture nonlinear distributions, but standard variants treat all minority samples equally, neglecting the importance of uncertain, boundary-region examples emphasized by heuristic methods like Borderline-SMOTE and ADASYN.

We propose Local Entropy-Guided Oversampling with a CVAE (LEO-CVAE), a generative oversampling framework that explicitly incorporates local uncertainty into both representation learning and data generation. To quantify uncertainty, we compute Shannon entropy over the class distribution in a sample's neighborhood: high entropy indicates greater class overlap, serving as a proxy for uncertainty. LEO-CVAE leverages this signal through two mechanisms: (i) a Local Entropy-Weighted Loss (LEWL) that emphasizes robust learning in uncertain regions, and (ii) an entropy-guided sampling strategy that concentrates generation in these informative, class-overlapping areas.

Applied to clinical genomics datasets (ADNI and TCGA lung cancer), LEO-CVAE consistently improves classifier performance, outperforming both traditional oversampling and generative baselines. These results highlight the value of uncertainty-aware generative oversampling for imbalanced learning in domains governed by complex nonlinear structures, such as omics data.
\end{abstract}

% --- MAIN BODY (TWO-COLUMN) ---
\begin{multicols}{2}
\section{Introduction}

The class imbalance problem, characterized by a severely skewed distribution of samples across classes, remains a critical challenge in machine learning \cite{ chen_survey_2024}.  Standard learning algorithms, optimized for overall accuracy, tend to develop a strong predictive bias towards the majority class \citep{Das2018}. Consequently, instances from the minority class, which are often of greatest interest in high-stakes domains like medical diagnosis, fraud detection, and industrial fault prediction, are frequently misclassified \citep{buda_systematic_2018, makki_experimental_2019, malhotra_empirical_2019}.

To mitigate this issue, a variety of techniques have been developed, broadly categorized into algorithm-level and data-level approaches \citep{Gao2025, Yang2024, buda_systematic_2018}. Data-level methods, which involve rebalancing the dataset prior to training, are particularly popular due to their model-agnostic nature \citep{chen_survey_2024, buda_systematic_2018}. Among these, oversampling techniques, which increase the representation of the minority class, are widely used \citep{azhar_investigation_2023}. The field has evolved from simple replication \citep{japkowicz_class_2000} to the landmark Synthetic Minority Over-sampling Technique (SMOTE) \citep{chawla_smote_2002}, which generates new, synthetic samples via linear interpolation between neighboring minority instances.

Despite its widespread adoption, SMOTE and its variants suffer from fundamental limitations. Their local, interpolative nature can generate noisy samples in regions of class overlap and fails to capture the global, often complex, distribution of the minority class \citep{Kamalov2025, Hong2024, tang_unbalanced_2025, wang_ctvae_2025}. Crucially, these methods typically ignore the rich distributional information present in the majority class, limiting the diversity and fidelity of the generated data \citep{ai_generative_2023}.

These shortcomings have motivated a paradigm shift towards deep generative models, which can learn global data distributions to generate novel and consistent samples \citep{Liu2007}. While Denoising Diffusion Models have achieved state-of-the-art performance in high-fidelity image synthesis \citep{Ho2020}, their application to tabular data remains a challenging area of research due to complex, non-Gaussian feature distributions \citep{liddm2025}. For tabular settings, they are also computationally intensive to train and sample from \citep{shi2025}. 

By contrast, Generative Adversarial Networks (GANs) \citep{goodfellow_generative_2014} and Variational Autoencoders (VAEs) \citep{Kingma2013} have been more widely explored for tabular data generation. GANs, however, often suffer from training instability \citep{sampath_survey_2021}, whereas VAEs provide a more stable and tractable alternative. Among them, the Conditional VAE (CVAE) \citep{Sohn2015} offers a principled framework for class-conditioned oversampling and has been successfully applied in imbalanced learning \citep{fajardo_oversampling_2021}. However, a key limitation of standard CVAEs is that they are agnostic to sample importance, treating all minority instances as equally informative. This uniform approach overlooks a critical insight from heuristic methods like Borderline-SMOTE \citep{han_borderline-smote_2005} and the Adaptive Synthetic Sampling (ADASYN) \citep{he_adasyn_2008}: not all samples are equally valuable for refining a classifier's decision boundary. Instances located deep within a class's feature space are less informative than those situated in regions of high class overlap, where the boundary is most ambiguous. These "borderline" or "hard-to-learn" samples are strategically crucial, as they provide the most challenging examples for a classifier to learn \citep{Japkowicz2002, he_adasyn_2008}.

Consequently, a significant gap remains in integrating this principle of uncertainty-awareness into the powerful distributional learning capabilities of a deep generative model. This gap is particularly pronounced in clinical genomics, which is the central focus of the current study. We hypothesize that the high dimensionality and intricate, nonlinear relationships in this domain cause the core assumption of linear interpolation used by SMOTE to break down, resulting in biologically implausible synthetic data \citep{ Blagus2012}. Furthermore, the inherent biological heterogeneity and the continuum-like nature of disease progression mean that class boundaries are rarely sharply defined, creating regions of high predictive uncertainty for a classifier. To leverage this insight, we turn to information theory to develop a formal measure of this local uncertainty. We quantify the degree of class overlap using Shannon entropy, the canonical measure of uncertainty, allowing us to identify high-entropy regions as a quantitative proxy for a sample’s 'hard-to-learn' status.

In essence, this combination of a complex, non-linear data manifold and inherently ambiguous class boundaries establishes clinical genomics as a uniquely challenging domain. This setting reveals the limitations of two key approaches: traditional oversampling methods, which struggle with complex, non-linear data, and standard generative models, which are agnostic to uncertain and hard-to-learn regions in the feature space. Consequently, it provides an ideal setting to validate a generative framework specifically engineered to target and learn from these high-uncertainty regions.

To this end, we propose the Local Entropy-Guided Oversampling with a CVAE (LEO-CVAE). We formalize the notion of a "hard-to-learn" or "uncertain" region using local Shannon entropy, a measure that quantifies the mixture of class labels within a sample's local neighborhood. The LEO-CVAE framework leverages this local entropy signal in two synergistic ways: first, it guides the CVAE's training process through a weighted loss function that prioritizes learning in the high-entropy regions; second, it directs the synthetic data generation by preferentially selecting high-entropy instances as seeds. By concentrating both the learning and generative processes on these ambiguous areas, LEO-CVAE directly reinforces the classifier's decision boundary where it is weakest. The primary contributions of this work are:
\begin{itemize}
    \item \textbf{A Novel Uncertainty Metric:} We introduce the Local Entropy Score (LES) to formally quantify sample-level uncertainty, identifying the most informative, class-overlapping regions for oversampling.
    \item \textbf{An Uncertainty-Aware Generative Framework:} We propose LEO-CVAE, which integrates the LES signal through two core components: a Local Entropy-Weighted Loss (LEWL) and an entropy-guided sampling strategy.
    \item \textbf{Empirical Validation:} We demonstrate the effectiveness of LEO-CVAE on challenging imbalanced clinical genomics datasets for both binary and multiclass classification through a systematic comparison against a suite of traditional and generative oversampling methods.
\end{itemize}

\section{Related Works}

\subsection{Traditional Oversampling and its Limitations}
Data-level approaches to class imbalance are dominated by oversampling techniques. The simplest method, Random Oversampling \citep{japkowicz_class_2000}, mitigates imbalance by duplicating minority class instances but is highly prone to overfitting. The Synthetic Minority Over-sampling Technique (SMOTE) \citep{chawla_smote_2002} provided a significant improvement by generating new samples via linear interpolation between a minority instance and its k-nearest minority neighbors. This approach has become a foundational baseline and has inspired a large family of variants \citep{Fernández2018, Li2025, Douzas2018, Douzas2019, Kunakorntum2020, Wang2025}. Borderline-SMOTE \citep{han_borderline-smote_2005}, for example, focuses generation on minority samples near the class boundary, which are deemed more critical for classification. The Adaptive Synthetic Sampling (ADASYN) \citep{he_adasyn_2008} method generates more synthetic data for minority samples that are harder to learn, based on the proportion of majority class instances in their local neighborhood.

Despite these advances, traditional methods share critical weaknesses. Their reliance on local, linear interpolation restricts the diversity of generated samples, often confining them to the convex hull of the original minority data \citep{w_dai_generative_2019, wang_ctvae_2025}. Furthermore, by focusing exclusively on minority class neighbors, they ignore the global data structure and the valuable information contained within the majority class, which can lead to the generation of noisy samples in regions of class overlap \citep{batista_study_2004, ai_generative_2023}.

\subsection{Deep Generative Models for Oversampling}
To overcome the limitations of local interpolation, research has shifted towards deep generative models, which learn an approximation of the entire data distribution, $p(x)$, enabling the generation of novel and globally consistent samples \citep{Liu2007}. The two most prominent architectures are Generative Adversarial Networks (GANs) \citep{goodfellow_generative_2014} and Variational Autoencoders (VAEs) \citep{Kingma2013}. While GANs can produce highly realistic samples, they often suffer from training instability and mode collapse \citep{sampath_survey_2021}. VAEs, by contrast, offer a stable, probabilistic framework for learning a smooth, continuous latent space from which to generate diverse data \citep{Kingma2013}.

The Conditional VAE (CVAE) \citep{Sohn2015} is a natural fit for oversampling, as it allows for targeted, class-specific data generation by conditioning the model on class labels. A CVAE is comprised of two networks: an encoder, $q_{\phi}(z|x, c)$, which learns to map data points into a probabilistic latent space conditioned on a class label c, and a decoder, $p_{\theta}(x|z, c)$, which learns to reconstruct data from that latent space. This provides a principled foundation for generating new minority samples by sampling from the latent space and decoding with the desired class label \citep{fajardo_oversampling_2021}.

\subsection{A Taxonomy of CVAE-Based Innovations}
Recent research has extended the CVAE framework for imbalanced learning along several distinct axes of innovation. Understanding these helps to precisely situate the contribution of LEO-CVAE.

\paragraph{Loss-Function Modification:} Drawing inspiration from discriminative learning, some approaches apply a focal loss to the CVAE's reconstruction objective to assign greater weight to samples with high reconstruction error \citep{t_y_lin_focal_2017}. This innovation modifies how the model learns from existing data.

\paragraph{Latent Space Structuring:} A second, powerful approach focuses on engineering a more discriminative latent space where class manifolds are well-separated. The Discriminative VAE (DVAE) \citep{guo_discriminative_2019} learns a latent mixture-of-Gaussians prior, which forces minority and majority samples into distinct clusters. This explicitly models the class boundary, enabling the generation of samples near this critical region. Similarly, the Contrastive Tabular VAE (CTVAE) \citep{wang_ctvae_2025} integrates a contrastive loss term into the training objective. This actively pulls latent representations of same-class samples together while pushing different-class samples apart, resulting in a highly structured latent space that improves the quality of generated samples. A key aspect of these methods is that after carefully structuring the latent space, they still typically sample randomly from the prior distribution for generation.

\paragraph{Knowledge Transfer:} For scenarios of extreme data scarcity, Majority-Guided VAE (MGVAE) \citep{ai_generative_2023} uses a transfer learning paradigm. It pre-trains on the abundant majority class to learn a robust feature representation and then fine-tunes on the few minority samples, effectively inheriting the diversity of the majority distribution. This approach excels at solving the specific problem of learning from few samples.

\vspace{1em}
Our proposed LEO-CVAE method contributes a distinct and novel perspective to this line of research. While the previously discussed methods focus on modifying the training loss, structuring the latent space, or implementing knowledge transfer, LEO-CVAE is the first to use local entropy as a direct signal to quantify sample-level uncertainty. Instead of treating all minority samples as equally important, LEO-CVAE identifies those in ambiguous, class-overlapped regions and intensifies both the model's learning and its generative process on these specific points. This allows LEO-CVAE to strategically reinforce the decision boundary where it is most contested, offering a new, uncertainty-aware approach to generative oversampling.

\section{Methods}

This section details our proposed oversampling method, Local Entropy-Guided Oversampling with a Conditional Variational Autoencoder (LEO-CVAE). We first provide the problem formulation, followed by an overview of the CVAE, which serves as our generative foundation. We then introduce our core contribution: the Local Entropy Score (LES), a metric for quantifying sample-level uncertainty to identify high-entropy regions within the feature space. Finally, we describe how LES is integrated into the CVAE through our two novel mechanisms: the Local Entropy-Weighted Loss (LEWL) for model training and the Entropy-Guided Sampling strategy for data generation.

\subsection{Problem Formulation}
Let $\mathcal{D} = \{(x_i, y_i)\}_{i=1}^N$ be a training dataset of $N$ samples, where $x_i \in \mathbb{R}^D$ is a $D$-dimensional feature vector and $y_i \in \{c_1, c_2, \dots, c_C\}$ is its corresponding class label. The dataset $\mathcal{D}$ is imbalanced if the class distribution is skewed, i.e., there exists a majority class $c_{maj}$ and a minority class $c_{min}$ such that the number of samples $N_{maj} \gg N_{min}$. The goal of oversampling is to generate a new set of synthetic minority samples, $\mathcal{D}_{syn}$, such that when combined with the original data ($\mathcal{D}' = \mathcal{D} \cup \mathcal{D}_{syn}$), the resulting dataset is balanced or near-balanced, leading to improved performance of a classifier trained on $\mathcal{D}'$.

\subsection{CVAE Foundation}
Our method is built upon a CVAE \citep{Sohn2015}, a generative model that learns a latent representation of data conditioned on class labels. A CVAE consists of two neural networks: an encoder and a decoder.

The encoder network, parameterized by $\phi$, learns to approximate the intractable true posterior distribution $p(z|x, c)$. It maps a data point x and its class condition c to the parameters of a diagonal Gaussian distribution, $q_{\phi}(z|x, c) = \mathcal{N}(z|\mu, \text{diag}(\sigma^2))$. The mean vector $\mu$ and variance vector $\sigma^2$ are the direct outputs of the encoder network.

The decoder network, parameterized by $\theta$, reconstructs the data by modeling the distribution $p_{\theta}(x|z, c)$. To generate a reconstructed sample $\hat{x}$, a latent vector z is first sampled from the encoder's output distribution using the reparameterization trick: $z = \mu + \epsilon \odot \sigma$, where $\epsilon \sim \mathcal{N}(0, I)$. This sample z is then concatenated with the one-hot class vector $c_{oh}$ and passed as input to the decoder.

The standard CVAE is trained by minimizing a loss function derived from the negative of the Evidence Lower Bound (ELBO):
\begin{equation}
\begin{split}
\mathcal{L}_{\text{CVAE}} ={} & -\mathbb{E}_{q_{\phi}(z|x,c)}[\log p_{\theta}(x|z, c)] \\
              & + \beta \cdot D_{KL}(q_{\phi}(z|x, c)||p(z|c))
\end{split}
\end{equation}

The first term is the reconstruction loss, which measures how well the model reconstructs the input data, and the second is the Kullback-Leibler (KL) divergence, which regularizes the latent space to follow a prior distribution $p(z|c)$.

\subsection{Local Entropy Score (LES)}
The core novelty of our method is the introduction of LES to guide the CVAE. LES quantifies the complexity of the feature space surrounding a given sample. For each sample $x_i$, we first identify its $k$ nearest neighbors, $\mathcal{N}_k(x_i)$. The neighborhood is determined using the standard Euclidean distance between feature vectors. We then compute the probability distribution of classes within this neighborhood. Let $k_j$ be the count of neighbors belonging to class $c_j$. The local probability of class $c_j$ is $p(c_j|x_i) = k_j/k$. The LES for sample $x_i$, denoted as $H(x_i)$, is then calculated using the Shannon entropy formula:
\begin{equation}
    H(x_i) = - \sum_{j=1}^{C} p(c_j|x_i) \log_2 p(c_j|x_i)
    \label{eq:entropy}
\end{equation}

For this calculation, we follow the standard convention that the contribution of any class $c_j$ with $p(c_j|x_i)=0$ is taken to be 0, as $\lim_{p\to0^+} p \log p = 0$.

The resulting score, which we denote $H_i = H(x_i)$, provides a quantitative measure of the heterogeneity of the sample's local feature space. The score ranges from a minimum of $0$, signifying a perfectly homogenous neighborhood where all neighbors belong to the same class, to a theoretical maximum of $\log_2(C)$, where $C$ is the total number of classes. This maximum value corresponds to a state of maximum entropy, representing the highest possible uncertainty where neighbors are uniformly distributed across all classes. A high LES, therefore, directly identifies a sample located in a complex, class-overlapping region of the feature space. This score becomes the critical signal for guiding both the learning and generation processes in our LEO-CVAE framework.

\subsection{The LEO-CVAE Method}
LEO-CVAE leverages the calculated local entropy scores in two critical stages: modifying the CVAE loss function to focus learning on high-entropy regions and guiding the generation of new synthetic samples.

\subsubsection{The Local Entropy-Weighted Loss (LEWL)}
To guide the CVAE's training, our primary innovation is a novel loss function, the LEWL. It instills uncertainty-awareness by modifying the CVAE's reconstruction component, while the standard KL divergence term is retained for its conventional regularization role. The complete LEWL objective is a composite function defined as:
\begin{equation}
    \mathcal{L}_{\text{LEWL}} = \mathcal{L}_{\text{W-Recon}} + \beta \cdot \mathcal{L}_{\text{KLD}}
    \label{eq:lewl_loss}
\end{equation}
Here, $\beta$ is a hyperparameter that weights the contribution of the Kullback-Leibler (KL) divergence term. We detail each component below.

\paragraph{Weighted Reconstruction Loss ($\mathcal{L}_{\text{W-Recon}}$):} The innovation of our training paradigm is captured in this component. We replace the CVAE's standard reconstruction error with a Weighted Mean Squared Error (MSE). This loss component strategically prioritizes samples that are most informative for learning a robust model: those belonging to minority classes and those located in complex, class-overlapping regions identified by a high LES. The loss for a single sample $x_i$ with class label $y_i$ and pre-calculated local entropy $H_i$ is defined as:
    \begin{equation}
        \mathcal{L}_{\text{W-Recon}, i} = w_{\text{class}}(y_i) \cdot w_{\text{entropy}}(H_i) \cdot \|x_i - \hat{x}_i\|^2_2
        \label{eq:w_recon_loss}
    \end{equation}
    The two weighting factors are:
    \begin{itemize}
        \item \textbf{Class-Imbalance Weight ($w_{\text{class}}$):} This weight is calculated from the inverse frequency of each class in the training data. By assigning a higher weight to samples from underrepresented classes, it compels the model to overcome the inherent bias of the imbalanced dataset.
        \item \textbf{Entropy-Focus Weight ($w_{\text{entropy}}$):} This factor directs the model's attention toward samples in ambiguous, high-entropy regions. It is defined as $w_{\text{entropy}}(H_i) = (1 + H_i)^{\gamma}$, where the hyperparameter $\gamma \geq 0$ controls the intensity of the focus. When $\gamma=0$, this guidance is inactive. For $\gamma > 0$, the model is more heavily penalized for failing to accurately reconstruct samples located in high-entropy regions, forcing it to learn a more discriminative representation of the decision boundary.
    \end{itemize}
The final reconstruction loss, $\mathcal{L}_{\text{W-Recon}}$, is the mean of these individually weighted losses calculated across all samples in a mini-batch.

\paragraph{KL Divergence Loss ($\mathcal{L}_{\text{KLD}}$):} To ensure the latent space remains well-structured, we include the standard KL Divergence Loss. This is the unmodified, conventional regularization term from the CVAE’s ELBO. It measures the divergence between the encoder's output distribution and a class-conditional prior and is defined as:
\begin{equation}
    \mathcal{L}_{\text{KLD}} = D_{\text{KL}}(q_{\phi}(z|x, c) || p(z|c))
\end{equation}
To counteract the common CVAE training issue of posterior collapse, where the KLD term vanishes and the decoder learns to ignore the latent code $z$, we enforce a minimum threshold on this loss component during optimization. This ensures that the latent variables continue to encode meaningful information throughout the training process.

\subsubsection{Entropy-Guided Sample Generation}
After training the LEO-CVAE, we use it to generate synthetic samples for each minority class until it reaches parity with the majority class. Applying the same core principle used in the LEWL, we focus the creation of new samples on high-entropy regions. The procedure for a given minority class $c_{min}$ is as follows:

\begin{enumerate}
    \item \textbf{Calculate Generation Count:} Determine the number of new samples to generate, $N_{gen} = N_{maj} - N_{min}$.
    \item \textbf{Select Seed Samples:} The seeds for generation are chosen from the original minority class instances in the training set. To prioritize the synthesis of new data in high-entropy regions, we employ a non-uniform selection strategy guided by LES. A sampling probability distribution, $P$, is established over the minority class samples where, instead of uniform selection, the probability of selecting a given sample $x_i$ is made proportional to its entropy-focus weight, the same transformation of its LES used in the LEWL:
    \begin{equation}
    \label{eq:entropy-guided generation}
    \begin{aligned}
    P(x_i) & \propto (1 + H_i)^\gamma, \\
    & \text{for all } (x_i, y_i) \text{ where } y_i = c_{\text{min}}
    \end{aligned}
    \end{equation}
    
     The hyperparameter $\gamma$ again controls how strongly this selection process favors samples in high-entropy regions. From this entropy-weighted distribution, $N_{\text{gen}}$ seed samples are drawn with replacement to initiate the data generation process. This ensures that samples residing in high-entropy neighborhoods are more likely to be chosen as templates for creating new synthetic data.

    \item \textbf{Generate Synthetic Data:} For each selected seed sample $x_{\text{seed}}$, a new synthetic sample $\hat{x}_{\text{new}}$ is generated. This is achieved by first encoding the seed to its latent distribution, then sampling a new latent vector $z_{\text{new}} \sim q_{\phi}(z|x_{\text{seed}}, c_{\text{min}})$ using the reparameterization trick. Finally, the resulting vector $z_{\text{new}}$ is passed to the decoder to produce the synthetic sample $\hat{x}_{\text{new}}$.
\end{enumerate}

This ensures that the synthetic data is generated around the most informative minority samples located in high-entropy regions, effectively reinforcing the decision boundary in contested regions of the feature space. The complete LEO-CVAE oversampling process is summarized in Algorithm \ref{alg:leocvae}.

\algrenewcommand\algorithmicrequire{\textbf{Input:}}
\algrenewcommand\algorithmicensure{\textbf{Output:}}
\begin{algorithm*}[t!]
    \caption{LEO-CVAE Framework}
    \label{alg:leocvae}
    \begin{algorithmic}[1]
        \Require Training data $\mathcal{D} = \{(x_i, y_i)\}_{i=1}^N$, $k$ for k-nearest neighbors, hyperparameters $\gamma, \beta$.
        \Ensure Resampled dataset $\mathcal{D}' = \mathcal{D} \cup \mathcal{D}_{syn}$.
        
        \Statex \textbf{Step 1: Quantifying Sample-Level Uncertainty}
        \For{each sample $x_i \in \mathcal{D}$}
            \State Find k-nearest neighbors $\mathcal{N}_k(x_i)$.
            \State Calculate local entropy score (LES) $H(x_i)$ using Eq.~\eqref{eq:entropy}.
        \EndFor
        
        \Statex \textbf{Step 2: Training with Entropy-Weighted Loss}
        \State Initialize LEO-CVAE model (Encoder$_\phi$, Decoder$_\theta$).
        \For{number of training epochs}
            \For{each mini-batch $\{(x_b, y_b, H_b)\}_{b=1}^B$}
                \State Compute $\mu_b, \log\sigma^2_b = \text{Encoder}_{\phi}([x_b, c_{b\_oh}])$.
                \State Sample $z_b \sim \mathcal{N}(\mu_b, \text{diag}(\sigma_b^2))$.
                \State Reconstruct $\hat{x}_b = \text{Decoder}_{\theta}([z_b, c_{b\_oh}])$.
                \State Calculate loss $\mathcal{L}_{\text{LEWL}}$ using Eq.~\eqref{eq:lewl_loss}.
                \State Update $\phi$ and $\theta$ via gradient descent.
            \EndFor
        \EndFor
        
        \Statex \textbf{Step 3: Entropy-Guided Generation}
        \State $\mathcal{D}_{syn} \leftarrow \emptyset$
        \State Identify minority classes $\mathcal{C}_{min}$ and majority count $N_{maj}$.
        \For{each class $c_j \in \mathcal{C}_{min}$}
            \State Let $\mathcal{D}_j$ be the set of samples in class $c_j$.
            \State Calculate sampling probabilities $P(x_i)$ for all $x_i \in \mathcal{D}_j$ using Eq.~\eqref{eq:entropy-guided generation}.
            \State $N_{gen} \leftarrow N_{maj} - |\mathcal{D}_j|$.
            \State Select $N_{gen}$ seed samples $\{x_{seed}\}$ from $\mathcal{D}_j$ with replacement using probabilities $P$.
            \State Generate $N_{gen}$ synthetic samples $\{\hat{x}_{new}\}$ from $\{x_{seed}\}$.
            \State $\mathcal{D}_{syn} \leftarrow \mathcal{D}_{syn} \cup \{(\hat{x}_{new}, c_j)\}$.
        \EndFor
        \State \Return Resampled dataset $\mathcal{D}' = \mathcal{D} \cup \mathcal{D}_{syn}$.
    \end{algorithmic}
\end{algorithm*}

\section{Experimental Setup}

Experimental validation was conducted on two challenging clinical genomics datasets: The Cancer Genome Atlas (TCGA) lung cancer and Alzheimer's Disease Neuroimaging Initiative (ADNI). These datasets were specifically selected, as they are characterized by the high dimensionality and complex, non-linear relationships inherent to genomic data, necessitating a powerful generative model. Moreover, the inherent biological heterogeneity among patients and the gradual progression of disease mean that class boundaries are not sharp, distinct lines. Instead, these factors create a continuum where samples from different classes intermingle, forming complex, overlapping decision boundaries. This results in the exact high-entropy regions that our entropy-guided ('LEO') mechanism is designed to target and resolve, providing a perfect testbed to rigorously evaluate our method's capacity to handle these real-world clinical data challenges.

\subsection{Datasets}
\paragraph{TCGA Lung Cancer Dataset:}
Gene expression (RNA-Seq) and corresponding clinical data for lung adenocarcinoma (LUAD) and lung squamous cell carcinoma (LUSC) were obtained from TCGA via the UCSC Xena \citep{Tomczak2015}. A total of 817 patient samples with complete records for both expression profiles and Progression-Free Survival (PFS) were included in this study. From an initial 17,738 gene expression features, the 64 most informative ones were selected using mutual information. The classification task was to predict patient PFS, which was categorized as 'short' ($\le$ 1 year) or 'long' ($>$ 1 year). The final cohort consisted of 147 'short' PFS samples (the minority class) and 670 'long' PFS samples (the majority class).

\paragraph{ADNI Dataset:}
Blood-based gene expression data were obtained from the ADNI database (\url{adni.loni.usc.edu}) for a multiclass classification task. ADNI, launched in 2003 as a public–private partnership led by Principal Investigator Michael W. Weiner, was designed to evaluate whether imaging, biological markers, and clinical/neuropsychological assessments could be combined to track the progression of mild cognitive impairment (MCI) and early Alzheimer’s disease (AD) \citep{Petersen2010}. Data from 744 participants were included in this analysis. Using a mutual information–based feature selection strategy, the initial 49,386 probe sets per sample were reduced to the 64 most informative features. The objective was to classify participants into Cognitively Normal (CN), MCI, or AD, with the dataset distributed as follows: 246 CN, 382 MCI, and 116 AD .

\subsection{Comparison Methods}
We benchmarked our proposed LEO-CVAE against a diverse suite of seven comparison methods. The specific configurations for each method are detailed in Supplement \ref{supp:baselines} to ensure full reproducibility.
\begin{enumerate}
    \item \textbf{No Oversampling}: The baseline performance of the classifier on the original, imbalanced data.
    \item \textbf{Random Oversampling \citep{japkowicz_class_2000}}: The simplest approach, which randomly duplicates samples from the minority class.
    \item \textbf{SMOTE \citep{chawla_smote_2002}}: The classic Synthetic Minority Over-sampling Technique.
    \item \textbf{Borderline-SMOTE \citep{han_borderline-smote_2005}}: An advanced SMOTE variant that focuses on the class boundary.
    \item \textbf{ADASYN \citep{he_adasyn_2008}}: An adaptive approach that generates more data for harder-to-learn minority samples.
    \item \textbf{Standard CVAE \citep{Sohn2015}}: A baseline CVAE model with an identical architecture to LEO-CVAE but using a standard, unweighted reconstruction loss.
    \item \textbf{CVAE with Focal Loss \citep{t_y_lin_focal_2017}}: A CVAE trained with a focal loss-inspired reconstruction objective to focus on samples with high reconstruction error.
\end{enumerate}

\subsection{Evaluation Protocol and Implementation Details}
All experiments followed a consistent and rigorous evaluation protocol to ensure a fair comparison.
\begin{itemize}
\item \textbf{Experimental Design:} We employed a 5-fold stratified cross-validation strategy. In each fold, oversampling methods were applied only to the training set. The validation set remained untouched to provide an unbiased estimate of performance. For reproducibility, all experiments were conducted with a fixed random seed of 42.

\item \textbf{Oversampling Models and Baselines:} To ensure a fair comparison, baseline models were configured with consistent parameters. The k-NN-based methods (i.e., SMOTE, Borderline-SMOTE, and ADASYN) used the same task-specific neighbor parameters. Similarly, our proposed LEO-CVAE and all other CVAE-based baselines shared an identical network architecture and core training parameters. Full details on the architectures and hyperparameters for all models are provided in Supplements \ref{supp:baselines} and \ref{supp:leocvae}.

\item \textbf{Downstream Classifier:} We used a Multi-Layer Perceptron (MLP) with a consistent architecture across all experiments to evaluate the quality of the data generated by each oversampling method. The MLP was trained using the Adam optimizer (learning rate: $1 \times 10^{-4}$, weight decay: $1 \times 10^{-3}$) with an early stopping patience of 20 epochs. Performance was monitored using the validation Area Under the Receiver Operating Characteristic Curve (AUC-ROC) for binary tasks and the micro-averaged AUC-ROC for multiclass tasks. A detailed description of the classifier's architecture is available in Supplement \ref{supp:MLPrarchitecture}.

\item \textbf{Evaluation Metrics:}
Model performance was evaluated using distinct sets of metrics for binary and multiclass classification tasks. For binary tasks, performance was assessed using the AUC-ROC, the Area Under the Precision-Recall Curve (AUPRC), and the F1-Score. For multiclass tasks, these same metrics were aggregated using both macro and micro averaging (macro/micro AUC-ROC, macro/micro AUPRC, and macro/micro F1-Score) to provide a more nuanced view of the model’s performance across all classes.
\end{itemize}

\begin{figure*}[t!]
    \centering
    \includegraphics[width=0.8\textwidth]{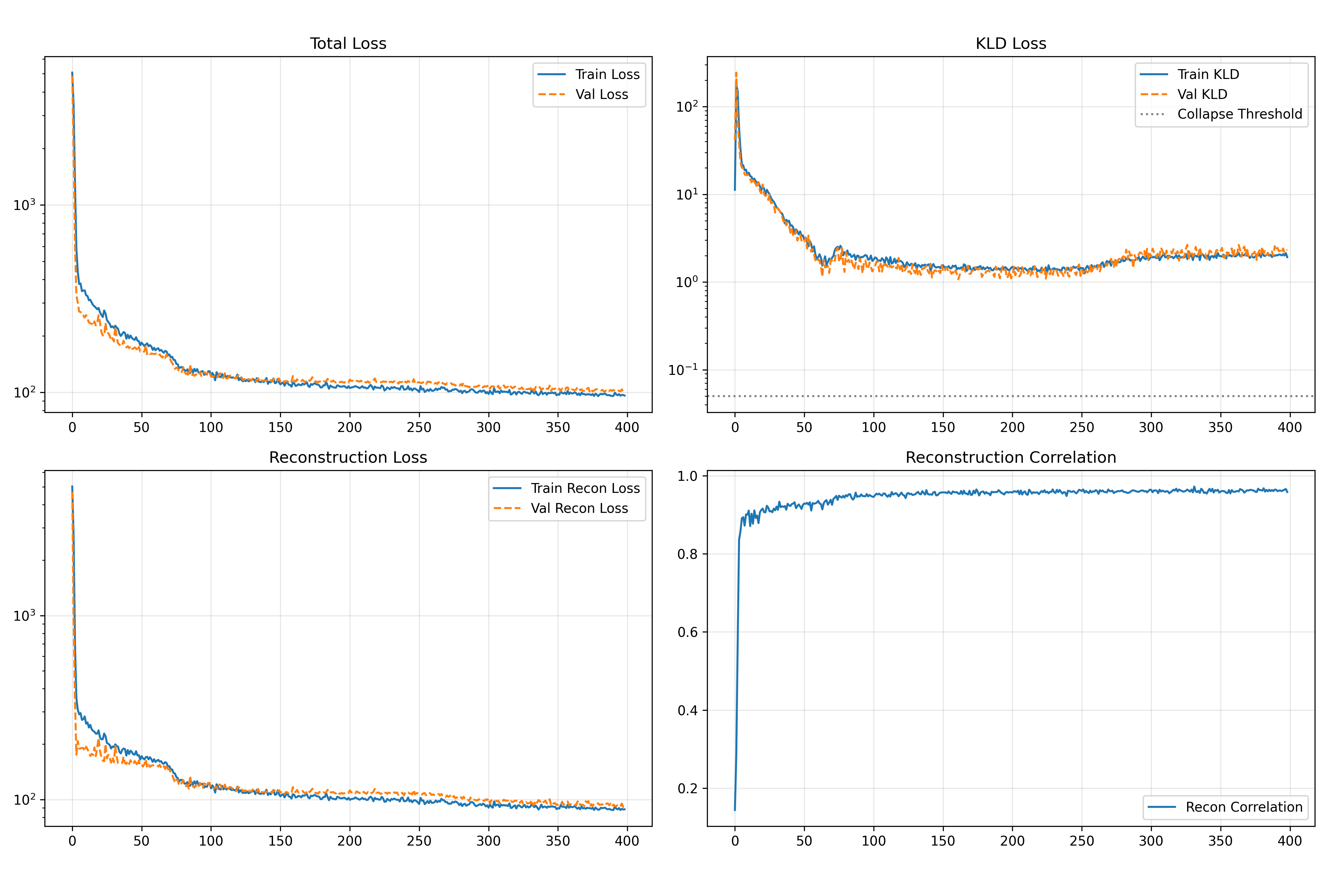}
    \caption{Training history of the LEO-CVAE model for a representative fold on the TCGA lung cancer dataset.}
    \label{fig:tcga_training_history}
\end{figure*}

\begin{lrbox}{\myboxtable}
\begin{tabular}{@{}lccc@{}}
\toprule
\textbf{Model} & \textbf{AUC-ROC} & \textbf{AUPRC} & \textbf{F1-Score} \\ \midrule
No Oversampling & 0.620 $ \pm $ 0.040 & 0.868 $ \pm $ 0.028 & \textbf{0.903} $ \pm $ 0.006 \\
Random Oversampling & 0.613 $ \pm $ 0.065 & 0.861 $ \pm $ 0.040 & 0.767 $ \pm $ 0.021 \\
SMOTE & 0.611 $ \pm $ 0.023 & 0.868 $ \pm $ 0.025 & 0.807 $ \pm $ 0.023 \\
Borderline-SMOTE & 0.630 $ \pm $ 0.044 & 0.874 $ \pm $ 0.022 & 0.797 $ \pm $ 0.026 \\
ADASYN & 0.617 $ \pm $ 0.052 & 0.869 $ \pm $ 0.031 & 0.786 $ \pm $ 0.026 \\
Standard CVAE & 0.614 $ \pm $ 0.074 & 0.869 $ \pm $ 0.040 & 0.883 $ \pm $ 0.015 \\
CVAE + Focal Loss & 0.645 $ \pm $ 0.043 & 0.885 $ \pm $ 0.016 & 0.871 $ \pm $ 0.028 \\
\textbf{LEO-CVAE} & \textbf{0.661} $ \pm $ 0.030 & \textbf{0.889} $ \pm $ 0.021 & 0.881 $ \pm $ 0.012 \\
\bottomrule
\end{tabular}
\end{lrbox}
\begin{table*}[t!]
\centering
\captionsetup{width=\wd\myboxtable, justification=raggedright, singlelinecheck=false}
\caption{Performance on the TCGA Lung Cancer Dataset}
\label{tab:main_tcga_results}
\usebox{\myboxtable}
\parbox{\wd\myboxtable}{\small \vspace{1ex}
\textit{Note:} Values are presented as mean $\pm$ standard deviation over 5 folds. The best result in each column is highlighted in bold.}
\end{table*}

% --- RESULTS ---
\section{Results}

\subsection{Performance on TCGA Lung Cancer Dataset (Binary Classification)}
The training dynamics of LEO-CVAE on the binary TCGA lung cancer dataset are illustrated for a representative fold in Figure \ref{fig:tcga_training_history}. The model exhibits stable convergence with no evidence of significant overfitting, as the validation and training loss curves closely track one another. The healthy stabilization of the KL Divergence and a final reconstruction correlation approaching 1.0 collectively affirm the successful and well-regularized training of the generative model.

The comparative classification results are presented in Table~\ref{tab:main_tcga_results}. LEO-CVAE achieved the highest AUC-ROC (0.661 $\pm$ 0.030) and AUPRC (0.889 $\pm$ 0.021), indicating improved ranking ability and better precision--recall trade-off for the minority class compared with all other methods. The No Oversampling baseline obtained the highest reported F1-score (0.903 $\pm$ 0.006).

Non-generative oversampling strategies, such as Borderline-SMOTE and ADASYN, showed modest increases in AUPRC compared to the baseline, but these were often accompanied by minimal or no gains in AUC-ROC, reflecting a common trade-off in which minority-class precision improves at the expense of overall discriminative ability. In contrast, CVAE-based approaches generally outperformed non-generative methods, with LEO-CVAE standing out as the only method to deliver notable, simultaneous gains in both AUC-ROC and AUPRC over the baseline.

\subsection{Performance on ADNI Dataset (Multiclass Classification)}
\begin{figure*}[t!]
    \centering
    \includegraphics[width=0.8\textwidth]{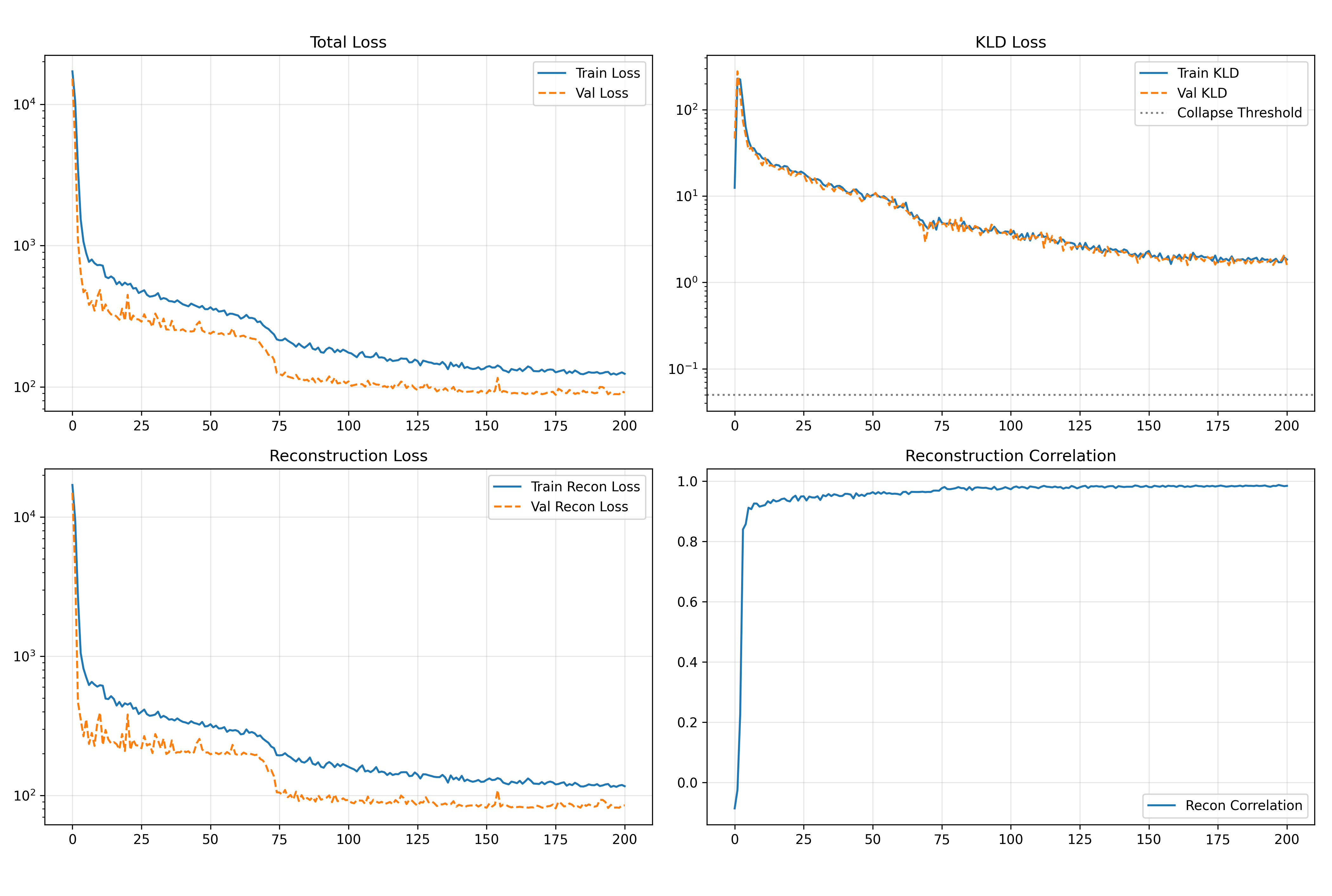}
    \caption{Training history of the LEO-CVAE model for a representative fold on the ADNI dataset.}
    \label{fig:adni_training_history}
\end{figure*}

The pattern of stable convergence was replicated on the multiclass ADNI dataset (Figure \ref{fig:adni_training_history}), establishing the reliability of the model's training process prior to performance evaluation.

For the multiclass ADNI dataset (Table~\ref{tab:main_adni_results}), the comparative results are more nuanced but again highlight the strengths of LEO-CVAE in achieving balanced performance across classes. The proposed model obtained the highest macro-averaged AUC-ROC (0.587 $\pm$ 0.025), macro-AUPRC (0.412 $\pm$ 0.015), and micro-AUPRC (0.500 $\pm$ 0.014), indicating improved per-class discrimination and a stronger precision--recall trade-off. The No Oversampling baseline performed strongly on micro-averaged metrics, such as micro-AUC-ROC (0.690 $\pm$ 0.017) and micro-F1 (0.500 $\pm$ 0.027), which weight performance by class size and are often dominated by populous classes. In contrast, LEO-CVAE's superior macro-averaged metrics (unweighted averages that treat all classes equally) highlight its ability to distribute performance gains more evenly, benefiting minority classes without sacrificing overall discrimination.

CVAE-based methods generally delivered stronger micro-metrics, likely due to generating a more diverse set of synthetic samples, but LEO-CVAE distinguished itself with a balanced macro/micro profile. 

\begin{table*}[t!]
\centering
\captionsetup{justification=raggedright, singlelinecheck=false}
\caption{Performance on the ADNI Dataset}
\label{tab:main_adni_results}
\resizebox{\textwidth}{!}{%
\begin{tabular}{@{}lcccccc@{}}
\toprule
& \multicolumn{2}{c}{\textbf{AUC-ROC}} & \multicolumn{2}{c}{\textbf{AUPRC}} & \multicolumn{2}{c}{\textbf{F1-Score}} \\
\cmidrule(lr){2-3} \cmidrule(lr){4-5} \cmidrule(lr){6-7}
\textbf{Model} & \textbf{Macro} & \textbf{Micro} & \textbf{Macro} & \textbf{Micro} & \textbf{Macro} & \textbf{Micro} \\ \midrule
No Oversampling & 0.570 $ \pm $ 0.033 & \textbf{0.690} $ \pm $ 0.017 & 0.398 $ \pm $ 0.022 & 0.492 $ \pm $ 0.021 & 0.311 $ \pm $ 0.026 & \textbf{0.500} $ \pm $ 0.027 \\
Random Oversampling & 0.564 $ \pm $ 0.029 & 0.588 $ \pm $ 0.020 & 0.394 $ \pm $ 0.027 & 0.393 $ \pm $ 0.018 & \textbf{0.381} $ \pm $ 0.012 & 0.402 $ \pm $ 0.020 \\
SMOTE & 0.561 $ \pm $ 0.041 & 0.606 $ \pm $ 0.030 & 0.403 $ \pm $ 0.035 & 0.416 $ \pm $ 0.035 & 0.377 $ \pm $ 0.043 & 0.410 $ \pm $ 0.040 \\
Borderline-SMOTE & 0.558 $ \pm $ 0.049 & 0.609 $ \pm $ 0.030 & 0.392 $ \pm $ 0.040 & 0.409 $ \pm $ 0.027 & 0.373 $ \pm $ 0.037 & 0.417 $ \pm $ 0.028 \\
ADASYN & 0.556 $ \pm $ 0.037 & 0.612 $ \pm $ 0.032 & 0.383 $ \pm $ 0.034 & 0.420 $ \pm $ 0.026 & 0.367 $ \pm $ 0.042 & 0.431 $ \pm $ 0.021 \\
Standard CVAE & 0.563 $ \pm $ 0.032 & 0.665 $ \pm $ 0.027 & 0.386 $ \pm $ 0.025 & 0.469 $ \pm $ 0.038 & 0.359 $ \pm $ 0.024 & 0.474 $ \pm $ 0.033 \\
CVAE + Focal Loss & 0.562 $ \pm $ 0.027 & 0.667 $ \pm $ 0.022 & 0.395 $ \pm $ 0.029 & 0.485 $ \pm $ 0.039 & 0.347 $ \pm $ 0.029 & 0.469 $ \pm $ 0.030 \\ 
\textbf{LEO-CVAE} & \textbf{0.587} $ \pm $ 0.025 & 0.683 $ \pm $ 0.013 & \textbf{0.412} $ \pm $ 0.015 & \textbf{0.500} $ \pm $ 0.014 & 0.375 $ \pm $ 0.043 & 0.484 $ \pm $ 0.020 \\
\bottomrule
\end{tabular}
}
\parbox{\textwidth}{\small \vspace{1ex}
\textit{Note:} Values are presented as mean $\pm$ standard deviation over 5 folds. The best result in each column is highlighted in bold.}
\end{table*}

\section{Ablation Study}

To rigorously evaluate the individual contributions of the core components within our proposed LEO-CVAE framework, a comprehensive ablation study was conducted. One or more of the model’s three main mechanisms were systematically disabled:
\begin{enumerate}
    \item Entropy-weighted loss 
    \item Entropy-guided sampling
    \item Inverse frequency class weighting
\end{enumerate}
The tested model variants are defined in Table \ref{tab:ablation_design}. All experiments followed the evaluation protocol outlined in Section 4.

\subsection{Results and Analysis}
The results for the binary TCGA lung cancer dataset are reported in Table~\ref{tab:tcga_ablation_results}, and for the multiclass ADNI dataset in Table~\ref{tab:adni_ablation_results}. Across both datasets, the full LEO-CVAE consistently achieved balanced, high performance, demonstrating the value of integrating all three mechanisms.
 
On the TCGA lung cancer dataset (Table~\ref{tab:tcga_ablation_results}), the full LEO-CVAE achieved the highest AUC-ROC ($0.661 \pm 0.030$) and AUPRC ($0.889 \pm 0.021$).  
Removing entropy-weighted loss (Ablation 1) or entropy-guided sampling (Ablation 2) reduced both metrics, confirming the utility of each entropy-based component. While the Standard CVAE achieved a high F1-score, this was accompanied by a comparatively low AUC-ROC.

% --- TABLE 3: ABLATION DESIGN (Full Width) ---
\begin{table*}[t!]
\centering
\captionsetup{justification=raggedright, singlelinecheck=false}
\caption{Ablation Study Experimental Design}
\label{tab:ablation_design}
\resizebox{\textwidth}{!}{%
\begin{tabular}{@{}lcccc@{}}
\toprule
\textbf{Model Variant} & \textbf{Entropy-Weighted Loss} & \textbf{Entropy-Guided Sampling} & \textbf{Class Weighting} & \textbf{Purpose} \\ \midrule
Full LEO-CVAE & \checkmark & \checkmark & \checkmark & The complete proposed model. \\
LEO-CVAE (w/o Ent-Weighted Loss) & \ding{55} & \checkmark & \checkmark & Isolates the effect of the entropy-weighted loss. \\
LEO-CVAE (w/o Ent-Guided Sampling) & \checkmark & \ding{55} & \checkmark & Isolates the effect of entropy-guided sampling. \\
LEO-CVAE (w/o Class Weights) & \checkmark & \checkmark & \ding{55} & Isolates the effect of class weighting. \\
CVAE + Class Weights & \ding{55} & \ding{55} & \checkmark & A simpler, non-entropy based model. \\
Standard CVAE & \ding{55} & \ding{55} & \ding{55} & The foundational generative baseline. \\ \bottomrule
\end{tabular}%
}
\end{table*}

\begin{lrbox}{\myboxtable}
\begin{tabular}{@{}lccc@{}}
\toprule
\textbf{Model Variant} & \textbf{AUC-ROC} & \textbf{AUPRC} & \textbf{F1-Score} \\ \midrule
\textbf{LEO-CVAE (Full Model)} & \textbf{0.661} $ \pm $ 0.030 & \textbf{0.889} $ \pm $ 0.021 & 0.881 $ \pm $ 0.012 \\
LEO-CVAE (w/o Entropy-Weighted Loss) & 0.600 $ \pm $ 0.054 & 0.863 $ \pm $ 0.036 & 0.861 $ \pm $ 0.025 \\
LEO-CVAE (w/o Entropy-Guided Sampling) & 0.631 $ \pm $ 0.032 & 0.875 $ \pm $ 0.025 & 0.861 $ \pm $ 0.026 \\
LEO-CVAE (w/o Class Weights) & 0.616 $ \pm $ 0.062 & 0.873 $ \pm $ 0.023 & 0.865 $ \pm $ 0.030 \\
CVAE + Class Weights & 0.617 $ \pm $ 0.026 & 0.865 $ \pm $ 0.025 & 0.868 $ \pm $ 0.027 \\
Standard CVAE & 0.614 $ \pm $ 0.074 & 0.869 $ \pm $ 0.040 & \textbf{0.883} $ \pm $ 0.015 \\
\bottomrule
\end{tabular}
\end{lrbox}
\begin{table*}[t!]
\centering
\captionsetup{width=\wd\myboxtable, justification=raggedright, singlelinecheck=false}
\caption{Ablation Study Results on the TCGA Lung Cancer Dataset}
\label{tab:tcga_ablation_results}
\usebox{\myboxtable}
\parbox{\wd\myboxtable}{\small \vspace{1ex} 
\textit{Note:} Values are presented as mean $\pm$ standard deviation over 5 folds. The best result in each column is highlighted in bold.}
\end{table*}
 
On the ADNI dataset (Table~\ref{tab:adni_ablation_results}), the full LEO-CVAE achieved the highest macro-averaged AUC ($0.587 \pm 0.025$), indicating balanced performance across all classes. Ablation 2 (no entropy-guided sampling) achieved the highest micro-averaged AUC ($0.686 \pm 0.019$) and macro F1-score ($0.379 \pm 0.013$), though differences with the full model were small and within standard deviations. This highlights the LEWL as an especially impactful component of the LEO-CVAE.

\begin{table*}[t!]
\centering
\captionsetup{justification=raggedright, singlelinecheck=false}
\caption{Ablation Study Results on the ADNI Dataset}
\label{tab:adni_ablation_results}
\resizebox{\textwidth}{!}{%
\begin{tabular}{@{}lcccccc@{}}
\toprule
& \multicolumn{2}{c}{\textbf{AUC-ROC}} & \multicolumn{2}{c}{\textbf{AUPRC}} & \multicolumn{2}{c}{\textbf{F1-Score}} \\
\cmidrule(lr){2-3} \cmidrule(lr){4-5} \cmidrule(lr){6-7}
\textbf{Model Variant} & \textbf{Macro} & \textbf{Micro} & \textbf{Macro} & \textbf{Micro} & \textbf{Macro} & \textbf{Micro} \\ \midrule
\textbf{LEO-CVAE (Full Model)} & \textbf{0.587} $ \pm $ 0.025 & 0.683 $ \pm $ 0.013 & \textbf{0.412} $ \pm $ 0.015 & \textbf{0.500} $ \pm $ 0.014 & 0.375 $ \pm $ 0.043 & 0.484 $ \pm $ 0.020 \\
LEO-CVAE (w/o Entropy-Weighted Loss) & 0.544 $ \pm $ 0.050 & 0.655 $ \pm $ 0.031 & 0.380 $ \pm $ 0.033 & 0.459 $ \pm $ 0.043 & 0.322 $ \pm $ 0.037 & 0.446 $ \pm $ 0.031 \\
LEO-CVAE (w/o Entropy-Guided Sampling) & 0.579 $ \pm $ 0.040 & \textbf{0.686} $ \pm $ 0.019 & \textbf{0.412} $ \pm $ 0.016 & 0.491 $ \pm $ 0.023 & \textbf{0.379} $ \pm $ 0.013 & \textbf{0.495} $ \pm $ 0.025 \\
LEO-CVAE (w/o Class Weights) & 0.533 $ \pm $ 0.019 & 0.646 $ \pm $ 0.013 & 0.377 $ \pm $ 0.021 & 0.445 $ \pm $ 0.018 & 0.350 $ \pm $ 0.034 & 0.458 $ \pm $ 0.030 \\
CVAE + Class Weights & 0.566 $ \pm $ 0.040 & 0.667 $ \pm $ 0.031 & 0.387 $ \pm $ 0.036 & 0.478 $ \pm $ 0.040 & 0.322 $ \pm $ 0.034 & 0.458 $ \pm $ 0.044 \\
Standard CVAE & 0.563 $ \pm $ 0.032 & 0.665 $ \pm $ 0.027 & 0.386 $ \pm $ 0.025 & 0.469 $ \pm $ 0.038 & 0.359 $ \pm $ 0.024 & 0.474 $ \pm $ 0.033 \\
\bottomrule
\end{tabular}
}
\parbox{\textwidth}{\small \vspace{1ex}
\textit{Note:} Values are presented as mean $\pm$ standard deviation over 5 folds. The best result in each column is highlighted in bold.}
\end{table*}

In summary, for the binary TCGA lung cancer dataset, all three components contribute to robust gains in AUC and AUPRC, with the entropy-based mechanisms having the largest impact. In the multiclass ADNI setting, the LEWL emerges as particularly effective, while removing entropy-guided sampling (Ablation 2) does not substantially harm performance and, in some cases, even slightly improves certain metrics. This validates the overall framework design by underscoring the power of the core entropy-weighted loss, while also highlighting that the optimal configuration of complementary components, such as class weighting and entropy-guided sampling, can be application-dependent.

\section{Conclusion}

In this work, we addressed the critical challenge of class imbalance in complex tabular data, with a focus on its application to clinical genomics data. Traditional oversampling methods, which rely on local, linear interpolation, are often ill-suited for such data. Their core assumption, that a straight line between two minority samples represents plausible data, frequently fails within the complex, non-linear manifolds characteristic of the biological processes underlying genomics data. While deep generative models like the Conditional Variational Autoencoder (CVAE) can capture these complex global distributions, a key limitation of standard CVAEs is that they learn the global distribution of a class by implicitly treating all training samples as equally informative. This uniform approach overlooks a critical insight: not all samples are equally valuable for refining a classifier’s decision boundary. Instances located deep within a class’s feature space are less informative than those situated in regions of high class overlap, where the boundary is most ambiguous. These "hard-to-learn" or "borderline" samples are strategically crucial, as they provide the most challenging examples for a classifier. While heuristic methods like Borderline-SMOTE and ADASYN pioneered the strategy of focusing on such instances, a significant gap remains in integrating a formal principle of uncertainty-awareness into a powerful distributional learning framework.

To bridge this gap, we introduce the Local Entropy-Guided Oversampling with a CVAE (LEO-CVAE), a novel generative oversampling framework that quantifies sample-level uncertainty using local Shannon entropy. By synergistically integrating this signal through a Local Entropy-Weighted Loss (LEWL) and an entropy-guided sampling strategy, LEO-CVAE focuses both its learning and generative processes on the most informative, class-overlapping regions of the feature space. Our empirical evaluation demonstrated that LEO-CVAE achieves superior and more balanced performance on challenging genomics datasets, delivering notable gains in AUC-ROC and AUPRC. The ablation study further revealed that the LEWL was the most impactful component, underscoring the benefit of compelling the model to learn a more robust representation of the contested decision boundary.

\subsection{Limitations and Future Directions}
This work lays the foundation for a new uncertainty-aware generative framework for imbalanced learning. We outline three promising avenues for future research based on the principles established in this study.

\subsubsection*{Applicability and Broader Evaluation}
Our study's focus on clinical genomics data was deliberate, as our framework is specifically designed for data with complex, non-linear characteristics. The principles of LEO-CVAE may also be well-suited for other high-dimensional omics data, such as proteomics and metabolomics, which share similar characteristics of complex, non-linear relationships. For simpler tabular data lacking these complex characteristics, traditional methods may still perform adequately. Future work should therefore extend this evaluation to a wider variety of tabular datasets, accompanied by formal statistical significance testing, to better characterize the regimes where different oversampling strategies are most effective.

\subsubsection*{Extensions to Alternative Generative Models}
The core concepts of entropy-guided learning and generation could be generalized to other data modalities, such as images, or adapted for other generative architectures like diffusion models and GANs. One particularly promising direction is adapting the LEO framework to diffusion-based synthesis. The denoising loss could be reweighted by the Local Entropy Score (LES) to counter majority-class gradient domination, while LES could also modulate class-conditional guidance during reverse diffusion, steering generation toward higher-quality minority samples. This integration of uncertainty-guided sampling with state-of-the-art synthesis offers a compelling path forward. Research also suggests that hybrid strategies combining traditional and generative models may yield further improvements \citep{wang_ctvae_2025}. 

\subsubsection*{Advancing the Uncertainty Metric}
The mechanism for calculating local entropy warrants deeper investigation. Our choice of k-NN with Euclidean distance was a direct extension of the principles in SMOTE \citep{chawla_smote_2002}, providing a robust baseline. However, the effectiveness of this uncertainty metric could be enhanced by exploring alternative distance metrics better suited for high-dimensional data, such as Manhattan distance, cosine similarity, and manifold-based distance, or by moving beyond distance-based methods with techniques like Kernel Density Estimation (KDE). A particularly novel direction would be to leverage the probability distributions from a pre-trained model to guide the entropy calculation, potentially creating a more nuanced, model-aware measure of uncertainty.

\section*{Acknowledgments}
Gene expression (RNA-Seq) and corresponding clinical data for lung adenocarcinoma (LUAD) and lung squamous cell carcinoma (LUSC) were obtained from TCGA via the UCSC Xena.

Data collection and sharing for this project was also funded by the Alzheimer's Disease Neuroimaging Initiative (ADNI) (National Institutes of Health Grant U01 AG024904) and DOD ADNI (Department of Defense award number W81XWH-12-2-0012). ADNI is funded by the National Institute on Aging, the National Institute of Biomedical Imaging and Bioengineering, and through generous contributions from the following: AbbVie, Alzheimer’s Association; Alzheimer’s Drug Discovery Foundation; Araclon Biotech; BioClinica, Inc.; Biogen; Bristol-Myers Squibb Company; CereSpir, Inc.; Cogstate; Eisai Inc.; Elan Pharmaceuticals, Inc.; Eli Lilly and Company; EuroImmun; F. Hoffmann-La Roche Ltd and its affiliated company Genentech, Inc.; Fujirebio; GE Healthcare; IXICO Ltd.; Janssen Alzheimer Immunotherapy Research \& Development, LLC.; Johnson \& Johnson Pharmaceutical Research \& Development LLC.; Lumosity; Lundbeck; Merck \& Co., Inc.; Meso Scale Diagnostics, LLC.; NeuroRx Research; Neurotrack Technologies; Novartis Pharmaceuticals Corporation; Pfizer Inc.; Piramal Imaging; Servier; Takeda Pharmaceutical Company; and Transition Therapeutics. The Canadian Institutes of Health Research is providing funds to support ADNI clinical sites in Canada. Private sector contributions are facilitated by the Foundation for the National Institutes of Health (\url{www.fnih.org}). The grantee organization is the Northern California Institute for Research and Education, and the study is coordinated by the Alzheimer’s Therapeutic Research Institute at the University of Southern California. ADNI data are disseminated by the Laboratory for Neuro Imaging at the University of Southern California.

\section*{Conflict of Interest}
The authors declare that the research was conducted in the absence of any commercial or financial relationships that could be construed as a potential conflict of interest.

\section*{Data Availability Statement}
The TCGA-LUAD and TCGA-LUSC datasets are publicly available via the UCSC Xena platform (\url{https://xenabrowser.net/}). The ADNI dataset is available to qualified researchers upon application through the Laboratory of Neuro Imaging (LONI) website (\url{https://adni.loni.usc.edu/}). Access to both datasets is subject to the data use agreements of their respective repositories. The source code for the LEO-CVAE framework is available at \url{https://github.com/Amirhossein-Zare/LEO-CVAE}.

% --- BIBLIOGRAPHY ---
%\bibliographystyle{plainnat}
%\bibliography{references}

\begin{thebibliography}{40}
\providecommand{\natexlab}[1]{#1}
\providecommand{\url}[1]{\texttt{#1}}
\expandafter\ifx\csname urlstyle\endcsname\relax
  \providecommand{\doi}[1]{doi: #1}\else
  \providecommand{\doi}{doi: \begingroup \urlstyle{rm}\Url}\fi

\bibitem[Ai et~al.(2023)Ai, Wang, He, Wen, Pan, and Xu]{ai_generative_2023}
Qingzhong Ai, Pengyun Wang, Lirong He, Liangjian Wen, Lujia Pan, and Zenglin Xu.
\newblock \emph{Generative {Oversampling} for {Imbalanced} {Data} via {Majority}-{Guided} {VAE}}.
\newblock 2023.

\bibitem[Azhar et~al.(2023)Azhar, Mohd~Pozi, Mohamed~Din, and Jatowt]{azhar_investigation_2023}
Nur Azhar, M.~S. Mohd~Pozi, Aniza Mohamed~Din, and Adam Jatowt.
\newblock An {Investigation} of {SMOTE} based {Methods} for {Imbalanced} {Datasets} with {Data} {Complexity} {Analysis}.
\newblock \emph{IEEE Transactions on Knowledge and Data Engineering}, 35:\penalty0 6651--6672, 2023.
\newblock \doi{10.1109/TKDE.2022.3179381}.

\bibitem[Batista et~al.(2004)Batista, Prati, and Monard]{batista_study_2004}
Gustavo Batista, Ronaldo Prati, and Maria-Carolina Monard.
\newblock A {Study} of the {Behavior} of {Several} {Methods} for {Balancing} machine {Learning} {Training} {Data}.
\newblock \emph{SIGKDD Explorations}, 6:\penalty0 20--29, 2004.
\newblock \doi{10.1145/1007730.1007735}.

\bibitem[Blagus and Lusa(2012)]{Blagus2012}
Rok Blagus and Lara Lusa.
\newblock Evaluation of smote for high-dimensional class-imbalanced microarray data.
\newblock volume~2, 12 2012.
\newblock \doi{10.1109/ICMLA.2012.183}.

\bibitem[Buda et~al.(2018)Buda, Maki, and Mazurowski]{buda_systematic_2018}
Mateusz Buda, Atsuto Maki, and Maciej~A. Mazurowski.
\newblock A systematic study of the class imbalance problem in convolutional neural networks.
\newblock \emph{Neural Networks}, 106:\penalty0 249--259, October 2018.
\newblock ISSN 0893-6080.
\newblock \doi{https://doi.org/10.1016/j.neunet.2018.07.011}.

\bibitem[Chawla et~al.(2002)Chawla, Bowyer, Hall, and Kegelmeyer]{chawla_smote_2002}
Nitesh Chawla, Kevin Bowyer, Lawrence Hall, and W.~Kegelmeyer.
\newblock {SMOTE}: {Synthetic} {Minority} {Over}-sampling {Technique}.
\newblock \emph{J. Artif. Intell. Res. (JAIR)}, 16:\penalty0 321--357, 2002.
\newblock \doi{10.1613/jair.953}.

\bibitem[Chen et~al.(2024)Chen, Yang, Yu, Shi, and Chen]{chen_survey_2024}
Wuxing Chen, Kaixiang Yang, Zhiwen Yu, Yifan Shi, and C.~Chen.
\newblock A survey on imbalanced learning: latest research, applications and future directions.
\newblock \emph{Artificial Intelligence Review}, 57, 2024.
\newblock \doi{10.1007/s10462-024-10759-6}.

\bibitem[Dai et~al.(2019)Dai, Ng, Severson, Huang, Anderson, and Stultz]{w_dai_generative_2019}
W.~Dai, K.~Ng, K.~Severson, W.~Huang, F.~Anderson, and C.~Stultz.
\newblock Generative {Oversampling} with a {Contrastive} {Variational} {Autoencoder}.
\newblock pages 101--109, November 2019.
\newblock ISBN 2374-8486.
\newblock \doi{10.1109/ICDM.2019.00020}.

\bibitem[Das et~al.(2018)Das, Datta, and Chaudhuri]{Das2018}
Swagatam Das, Shounak Datta, and Bidyut Chaudhuri.
\newblock Handling data irregularities in classification: Foundations, trends, and future challenges.
\newblock \emph{Pattern Recognition}, 81, 03 2018.
\newblock \doi{10.1016/j.patcog.2018.03.008}.

\bibitem[Douzas and Bação(2019)]{Douzas2019}
Georgios Douzas and Fernando Bação.
\newblock Geometric smote a geometrically enhanced drop-in replacement for smote.
\newblock \emph{Information Sciences}, 501, 06 2019.
\newblock \doi{10.1016/j.ins.2019.06.007}.

\bibitem[Douzas et~al.(2018)Douzas, Bação, and Last]{Douzas2018}
Georgios Douzas, Fernando Bação, and Felix Last.
\newblock Improving imbalanced learning through a heuristic oversampling method based on k-means and smote.
\newblock \emph{Information Sciences}, 465, 06 2018.
\newblock \doi{10.1016/j.ins.2018.06.056}.

\bibitem[Fajardo et~al.(2021)Fajardo, Findlay, Jaiswal, Yin, Houmanfar, Xie, Liang, She, and Emerson]{fajardo_oversampling_2021}
Val~Andrei Fajardo, David Findlay, Charu Jaiswal, Xinshang Yin, Roshanak Houmanfar, Honglei Xie, Jiaxi Liang, Xichen She, and D.~B. Emerson.
\newblock On oversampling imbalanced data with deep conditional generative models.
\newblock \emph{Expert Systems with Applications}, 169, 2021.
\newblock ISSN 09574174.
\newblock \doi{10.1016/j.eswa.2020.114463}.

\bibitem[Fernández et~al.(2018)Fernández, Garcia, Herrera, and Chawla]{Fernández2018}
Alberto Fernández, Salvador Garcia, Francisco Herrera, and Nitesh Chawla.
\newblock Smote for learning from imbalanced data: Progress and challenges, marking the 15-year anniversary.
\newblock \emph{Journal of Artificial Intelligence Research}, 61:\penalty0 863--905, 04 2018.
\newblock \doi{10.1613/jair.1.11192}.

\bibitem[Gao et~al.(2025)Gao, Xie, Zhang, Wang, He, Yin, and Zhang]{Gao2025}
Xinyi Gao, Dongting Xie, Yihang Zhang, Zhengren Wang, Conghui He, Hongzhi Yin, and Wentao Zhang.
\newblock A comprehensive survey on imbalanced data learning.
\newblock 02 2025.
\newblock \doi{10.48550/arXiv.2502.08960}.

\bibitem[Goodfellow et~al.(2014)Goodfellow, Pouget-Abadie, Mirza, Xu, Warde-Farley, Ozair, Courville, and Bengio]{goodfellow_generative_2014}
Ian Goodfellow, Jean Pouget-Abadie, Mehdi Mirza, Bing Xu, David Warde-Farley, Sherjil Ozair, Aaron Courville, and Y.~Bengio.
\newblock Generative {Adversarial} {Networks}.
\newblock \emph{Advances in Neural Information Processing Systems}, 3, 2014.
\newblock \doi{10.1145/3422622}.

\bibitem[Guo et~al.(2019)Guo, Zhu, Wang, and Chen]{guo_discriminative_2019}
Ting Guo, Xingquan Zhu, Yang Wang, and Fang Chen.
\newblock \emph{Discriminative {Sample} {Generation} for {Deep} {Imbalanced} {Learning}}.
\newblock 2019.

\bibitem[Haibo et~al.(2008)Haibo, Yang, Garcia, and Shutao]{he_adasyn_2008}
He~Haibo, Bai Yang, E.~A. Garcia, and Li~Shutao.
\newblock {ADASYN}: {Adaptive} synthetic sampling approach for imbalanced learning.
\newblock pages 1322--1328, June 2008.
\newblock ISBN 2161-4407.
\newblock \doi{10.1109/IJCNN.2008.4633969}.

\bibitem[Han et~al.(2005)Han, Wang, and Mao]{han_borderline-smote_2005}
Hui Han, Wen-Yuan Wang, and Bing-Huan Mao.
\newblock Borderline-{SMOTE}: {A} {New} {Over}-{Sampling} {Method} in {Imbalanced} {Data} {Sets} {Learning}.
\newblock pages 878--887. Springer Berlin Heidelberg, 2005.
\newblock ISBN 978-3-540-31902-3.

\bibitem[Ho et~al.(2020)Ho, Jain, and Abbeel]{Ho2020}
Jonathan Ho, Ajay Jain, and Pieter Abbeel.
\newblock Denoising diffusion probabilistic models, 06 2020.

\bibitem[Hong et~al.(2024)Hong, An, and Jeon]{Hong2024}
Sungchul Hong, Seunghwan An, and Jong-June Jeon.
\newblock Improving smote via fusing conditional vae for data-adaptive noise filtering.
\newblock 05 2024.
\newblock \doi{10.48550/arXiv.2405.19757}.

\bibitem[Japkowicz(2000)]{japkowicz_class_2000}
Nathalie Japkowicz.
\newblock The {Class} {Imbalance} {Problem}: {Significance} and {Strategies}.
\newblock \emph{Proceedings of the 2000 International Conference on Artificial Intelligence ICAI}, 2000.

\bibitem[Japkowicz and Stephen(2002)]{Japkowicz2002}
Nathalie Japkowicz and Shaju Stephen.
\newblock The class imbalance problem: A systematic study.
\newblock \emph{Intell. Data Anal.}, 6:\penalty0 429--449, 11 2002.
\newblock \doi{10.3233/IDA-2002-6504}.

\bibitem[Kamalov et~al.(2025)Kamalov, Choutri, and Atiya]{Kamalov2025}
Firuz Kamalov, Salah Choutri, and Amir Atiya.
\newblock Analytical formulation of synthetic minority oversampling technique (smote) for imbalanced learning.
\newblock \emph{Gulf Journal of Mathematics}, 19:\penalty0 400--415, 01 2025.
\newblock \doi{10.56947/gjom.v19i1.2639}.

\bibitem[Kingma and Welling(2013)]{Kingma2013}
Diederik Kingma and Max Welling.
\newblock Auto-encoding variational bayes.
\newblock \emph{ICLR}, 12 2013.

\bibitem[Kunakorntum et~al.(2020)Kunakorntum, Hinthong, and Phunchongharn]{Kunakorntum2020}
Intouch Kunakorntum, Woranich Hinthong, and Phond Phunchongharn.
\newblock A synthetic minority based on probabilistic distribution (symprod) oversampling for imbalanced datasets.
\newblock \emph{IEEE Access}, PP:\penalty0 1--1, 06 2020.
\newblock \doi{10.1109/ACCESS.2020.3003346}.

\bibitem[Li et~al.(2025{\natexlab{a}})Li, Yang, Song, Duan, and Ren]{Li2025}
Ying Li, Yali Yang, Peihua Song, Lian Duan, and Rui Ren.
\newblock An improved smote algorithm for enhanced imbalanced data classification by expanding sample generation space.
\newblock \emph{Scientific Reports}, 15, 07 2025{\natexlab{a}}.
\newblock \doi{10.1038/s41598-025-09506-w}.

\bibitem[Li et~al.(2025{\natexlab{b}})Li, Huang, Yang, Shi, Yang, van Stein, Bäck, and van Leeuwen]{liddm2025}
Zhong Li, Qi~Huang, Lincen Yang, Jiayang Shi, Zhao Yang, Niki van Stein, Thomas Bäck, and Matthijs van Leeuwen.
\newblock Diffusion models for tabular data: Challenges, current progress, and future directions, 2025{\natexlab{b}}.
\newblock URL \url{https://arxiv.org/abs/2502.17119}.

\bibitem[Lin et~al.(2017)Lin, Goyal, Girshick, He, and Dollár]{t_y_lin_focal_2017}
T.~Y. Lin, P.~Goyal, R.~Girshick, K.~He, and P.~Dollár.
\newblock Focal {Loss} for {Dense} {Object} {Detection}.
\newblock pages 2999--3007, October 2017.
\newblock ISBN 2380-7504.
\newblock \doi{10.1109/ICCV.2017.324}.

\bibitem[Liu et~al.(2007)Liu, Ghosh, and Martin]{Liu2007}
Alexander Liu, Joydeep Ghosh, and Cheryl Martin.
\newblock Generative oversampling for mining imbalanced datasets.
\newblock pages 66--72, 01 2007.

\bibitem[Makki et~al.(2019)Makki, Assaghir, Taher, Haque, Hacid, and Zeineddine]{makki_experimental_2019}
Sara Makki, Zainab Assaghir, Yehia Taher, Rafiqul Haque, Mohand-Said Hacid, and Hassan Zeineddine.
\newblock An {Experimental} {Study} {With} {Imbalanced} {Classification} {Approaches} for {Credit} {Card} {Fraud} {Detection}.
\newblock \emph{IEEE Access}, PP:\penalty0 1--1, 2019.
\newblock \doi{10.1109/ACCESS.2019.2927266}.

\bibitem[Malhotra and Kamal(2019)]{malhotra_empirical_2019}
Ruchika Malhotra and Shine Kamal.
\newblock An {Empirical} {Study} to {Investigate} {Oversampling} {Methods} for {Improving} {Software} {Defect} {Prediction} using {Imbalanced} {Data}.
\newblock \emph{Neurocomputing}, 343, 2019.
\newblock \doi{10.1016/j.neucom.2018.04.090}.

\bibitem[Petersen et~al.(2010)Petersen, Aisen, Beckett, Donohue, Gamst, Harvey, Jack, Jagust, Shaw, Toga, Trojanowski, and Weiner]{Petersen2010}
Ronald Petersen, P.S. Aisen, L.A. Beckett, Michael Donohue, A.C. Gamst, D.J. Harvey, Clifford Jack, W.J. Jagust, Leslie Shaw, A.W. Toga, J.Q. Trojanowski, and Michael Weiner.
\newblock Alzheimer's disease neuroimaging initiative (adni): Clinical characterization.
\newblock \emph{Neurology}, 74:\penalty0 201--9, 01 2010.
\newblock \doi{10.1212/WNL.0b013e3181cb3e25}.

\bibitem[Sampath et~al.(2021)Sampath, Maurtua, Aguilar~Martín, and Gutierrez]{sampath_survey_2021}
V.~Sampath, I.~Maurtua, J.~J. Aguilar~Martín, and A.~Gutierrez.
\newblock A survey on generative adversarial networks for imbalance problems in computer vision tasks.
\newblock \emph{J Big Data}, 8\penalty0 (1):\penalty0 27, 2021.
\newblock ISSN 2196-1115 (Print) 2196-1115.
\newblock \doi{10.1186/s40537-021-00414-0}.

\bibitem[Shi et~al.(2025)Shi, Zhang, Tong, and Xu]{shi2025}
Dingyuan Shi, Lulu Zhang, Yongxin Tong, and Ke~Xu.
\newblock Understanding and mitigating the high computational cost in path data diffusion, 2025.
\newblock URL \url{https://arxiv.org/abs/2502.00725}.

\bibitem[Sohn et~al.(2015)Sohn, Lee, and Yan]{Sohn2015}
Kihyuk Sohn, Honglak Lee, and Xinchen Yan.
\newblock Learning structured output representation using deep conditional generative models.
\newblock In C.~Cortes, N.~Lawrence, D.~Lee, M.~Sugiyama, and R.~Garnett, editors, \emph{Advances in Neural Information Processing Systems}, volume~28. Curran Associates, Inc., 2015.
\newblock URL \url{https://proceedings.neurips.cc/paper_files/paper/2015/file/8d55a249e6baa5c06772297520da2051-Paper.pdf}.

\bibitem[Tang et~al.(2025)Tang, Li, Sun, Wen, and Guo]{tang_unbalanced_2025}
Xi~Tang, Wenhai Li, Weichao Sun, Tianzhu Wen, and Kai Guo.
\newblock \emph{Unbalanced {Data} {Oversampling} {Method} {Based} on {Improved} {VAE}-{CGAN}}.
\newblock 2025.

\bibitem[Tomczak et~al.(2015)Tomczak, Czerwinska, and Wiznerowicz]{Tomczak2015}
Katarzyna Tomczak, Patrycja Czerwinska, and Maciej Wiznerowicz.
\newblock The cancer genome atlas (tcga): An immeasurable source of knowledge.
\newblock \emph{Contemp Oncol (Pozn)}, 19:\penalty0 A68--A77, 01 2015.

\bibitem[Wang et~al.(2025{\natexlab{a}})Wang, Le, Duong, Van, and Nguyen]{wang_ctvae_2025}
Alex~X. Wang, Minh~Quang Le, Huu-Thanh Duong, Bay~Nguyen Van, and Binh~P. Nguyen.
\newblock {CTVAE}: {Contrastive} {Tabular} {Variational} {Autoencoder} for imbalance data.
\newblock \emph{Knowledge and Information Systems}, 67\penalty0 (6):\penalty0 5335--5354, 2025{\natexlab{a}}.
\newblock ISSN 0219-1377 0219-3116.
\newblock \doi{10.1007/s10115-025-02377-7}.

\bibitem[Wang et~al.(2025{\natexlab{b}})Wang, Le, Nguyen~Trung, and Nguyen]{Wang2025}
Alex~Xing Wang, Viet-Tuan Le, Hau Nguyen~Trung, and Binh Nguyen.
\newblock Addressing imbalance in health data: Synthetic minority oversampling using deep learning.
\newblock \emph{Computers in biology and medicine}, 188:\penalty0 109830, 02 2025{\natexlab{b}}.
\newblock \doi{10.1016/j.compbiomed.2025.109830}.

\bibitem[Yang et~al.(2024)Yang, Yu, Chen, Liang, and Chen]{Yang2024}
Kaixiang Yang, Zhiwen Yu, Wuxing Chen, Zefeng Liang, and C.~Chen.
\newblock Solving the imbalanced problem by metric learning and oversampling.
\newblock \emph{IEEE Transactions on Knowledge and Data Engineering}, PP:\penalty0 1--14, 12 2024.
\newblock \doi{10.1109/TKDE.2024.3419834}.

\end{thebibliography}

\end{multicols}
% The supplementary materials start below.
\newpage
\onecolumn
\appendix
% --- Renumber sections, subsections, and tables for Supplementary Materials ---
\renewcommand{\thesection}{s}
\renewcommand{\thesubsection}{\thesection.\arabic{subsection}}
\renewcommand{\thetable}{\thesection\arabic{table}}
\setcounter{table}{0}

\section{Model Architectures and Hyperparameters}
This supplement provides detailed information on the network architectures and hyperparameter settings used in our experiments to ensure full reproducibility.

\subsection{Baseline Model Configurations}
\label{supp:baselines}
Non-generative baselines were implemented using the \texttt{imbalanced-learn} Python library. All CVAE-based baselines use the identical network architecture as LEO-CVAE (Supplement \ref{supp:leocvae}). Specific configurations are in Table \ref{tab:baseline_params}.

\begin{table}[H]
\centering
\caption{Baseline Model Hyperparameters}
\label{tab:baseline_params}
\begin{tabularx}{\textwidth}{@{} l c >{\RaggedRight}X >{\RaggedRight}X @{}}
\toprule
\textbf{Model} & \textbf{Library/Base} & \textbf{Key Parameter} & \textbf{Value} \\ \midrule
\textbf{SMOTE} & \texttt{imbalanced-learn} & `k\_neighbors` & 7 (Binary) / 15 (Multiclass) \\ \addlinespace
\textbf{Borderline-SMOTE} & \texttt{imbalanced-learn} & `kind` & 'borderline-1' \\
 & & `k\_neighbors` & 7 (Binary) / 15 (Multiclass) \\
 & & `m\_neighbors` & 7 (Binary) / 15 (Multiclass) \\ \addlinespace
\textbf{ADASYN} & \texttt{imbalanced-learn} & `n\_neighbors` & 7 (Binary) / 15 (Multiclass) \\ \addlinespace
\textbf{Standard CVAE} & Our implementation & KLD Weight ($\beta$) & 1.0 \\
 & & Minimum KLD & 0.1 \\ \addlinespace
\textbf{CVAE with Focal Loss} & Our implementation & KLD Weight ($\beta$) & 1.0 \\
 & & Minimum KLD & 0.1 \\
 & & Focusing Param. ($\gamma$) & 1.0 \\ \bottomrule
\end{tabularx}
\end{table}

\subsection{LEO-CVAE Architecture and Hyperparameters}
\label{supp:leocvae}
Our proposed LEO-CVAE model and all CVAE-based baselines share the identical network architecture detailed below, with an input dimension of 64.
\begin{itemize}
    \item \textbf{Encoder Architecture}:
    \begin{enumerate}
        \item Input: Concatenation of feature vector ($D_{in}=64$) and one-hot class label ($C$), size = $64 + C$.
        \item `Linear`($64 + C \rightarrow 64$) $\rightarrow$ `LeakyReLU`(0.2) $\rightarrow$ `Dropout`($p=0.1$).
        \item `Linear`(64 $\rightarrow$ 32) $\rightarrow$ `LeakyReLU`(0.2) $\rightarrow$ `Dropout`($p=0.1$).
        \item Two parallel `Linear` output heads from the 32-neuron layer:
        \begin{itemize}
            \item `Linear`(32 $\rightarrow$ 16) for the latent mean ($\mu$).
            \item `Linear`(32 $\rightarrow$ 16) for the latent log-variance ($\log\sigma^2$).
        \end{itemize}
    \end{enumerate}
    \item \textbf{Decoder Architecture}:
    \begin{enumerate}
        \item Input: Concatenation of latent vector ($D_z=16$) and one-hot class label ($C$), size = $16 + C$.
        \item `Linear`($16 + C \rightarrow 32$) $\rightarrow$ `LeakyReLU`(0.2) $\rightarrow$ `Dropout`($p=0.1$).
        \item `Linear`(32 $\rightarrow$ 64) $\rightarrow$ `LeakyReLU`(0.2) $\rightarrow$ `Dropout`($p=0.1$).
        \item `Linear` output layer mapping from $64 \rightarrow 64$ to reconstruct the original feature vector.
    \end{enumerate}
\end{itemize}
The hyperparameters for training LEO-CVAE are detailed in Table \ref{tab:leocvae_params}.

\begin{table}[H]
\centering
\caption{LEO-CVAE Training Hyperparameters}
\label{tab:leocvae_params}
\begin{tabularx}{\textwidth}{@{} l c >{\RaggedRight}X >{\RaggedRight}X @{}}
\toprule
\textbf{Hyperparameter} & \textbf{Symbol} & \textbf{Value} & \textbf{Description} \\ \midrule
Optimizer & - & Adam & - \\
Learning Rate & - & $1 \times 10^{-3}$ & - \\
Weight Decay & - & $1 \times 10^{-5}$ & - \\
Batch Size & - & 32 & - \\
Max Epochs & - & 500 & - \\
Early Stopping Patience & - & 25 & - \\
Gradient Clip Norm & - & 1.0 & Maximum norm for gradient clipping. \\
Latent Dimension & $D_z$ & 16 & Dimensionality of the latent space. \\
\ Entropy k-nearest neighbors & $k$ & \ 7 (Binary) / 15 (Multiclass) & Neighbors for local entropy calculation. \\
\ Entropy Focus & $\gamma$ & \ 0.5 (Binary) / 2.5 (Multiclass) & Weighting for high-entropy samples. \\
KLD Weight & $\beta$ & 4.0 & Weight of the KL divergence term. \\
Minimum KLD & - & 0.1 & Floor to prevent posterior collapse. \\
\bottomrule
\end{tabularx}
\end{table}

\subsection{MLP Classifier Architecture}
\label{supp:MLPrarchitecture}
The Multi-Layer Perceptron (MLP) used as the downstream classifier has an input dimension of 64 and the following single-hidden-layer architecture:
\begin{itemize}
    \item \textbf{Hidden Layer:} `Linear` layer ($64 \rightarrow 32$) $\rightarrow$ `BatchNorm1d` $\rightarrow$ `ReLU` $\rightarrow$ `Dropout` ($p = 0.5$).
    \item \textbf{Output Layer:} `Linear` layer mapping from $32 \rightarrow 1$ (Binary) or $32 \rightarrow 3$ (Multiclass).
\end{itemize}
The MLP was trained using the hyperparameters listed in Table \ref{tab:mlp_params}.

\begin{table}[H]
\centering
\caption{MLP Classifier Training Hyperparameters}
\label{tab:mlp_params}
\begin{tabular}{@{}ll@{}}
\toprule
\textbf{Hyperparameter}        & \textbf{Value}                         \\ \midrule
Optimizer                      & Adam                                   \\
Learning Rate                  & $1 \times 10^{-4}$                     \\
Weight Decay                   & $1 \times 10^{-3}$                     \\
Batch Size                     & 32                                     \\
Max Epochs                     & 200                                    \\ \addlinespace
LR Scheduler                   & ReduceLROnPlateau                      \\
LR Scheduler Patience          & 5                                      \\
LR Scheduler Factor            & 0.7                                    \\ \addlinespace
Early Stopping Patience        & 20                                     \\
Early Stopping Metric          & AUC (Binary) / AUC Micro (Multiclass)  \\ \addlinespace
Gradient Clip Norm             & 0.5  
\\
Label Smoothing                & 0.1
\\ \bottomrule
\end{tabular}
\end{table}

\end{document}